%% file: main.tex
\documentclass[10pt,twocolumn,letterpaper]{article}

\usepackage[pagenumbers]{cvpr} 

\usepackage{mathrsfs}
\usepackage{cuted}
\usepackage{xcolor}

\usepackage{graphicx}
\usepackage{placeins}  
\usepackage{capt-of}  
\usepackage{multirow}

\input{preamble}

\definecolor{cvprblue}{rgb}{0.21,0.49,0.74}
\usepackage[pagebackref,breaklinks,colorlinks,allcolors=cvprblue]{hyperref}

\def\paperID{10883} 
\def\confName{CVPR}
\def\confYear{2026}

\title{Rethinking Structure Preservation in Text-Guided Image Editing with Visual Autoregressive Models}

\author{
Tao Xia$^{1}$, Jiawei Liu$^{2}$, Yukun Zhang$^{1}$, Ting Liu$^{3}$, Wei Wang$^{2}$, Lei Zhang$^{1,\star}$\\
$^1$Beijing Institute of Technology, Beijing, China \\
$^2$Shenyang Institute of Automation, CAS, Shenyang, China \\
$^3$Meitu Inc, MTLab, Beijing, China \\
{\tt\small $\star$leizhang@bit.edu.cn}
}
\begin{document}
\maketitle

\begin{strip}
  \centering
  \includegraphics[width=\linewidth]{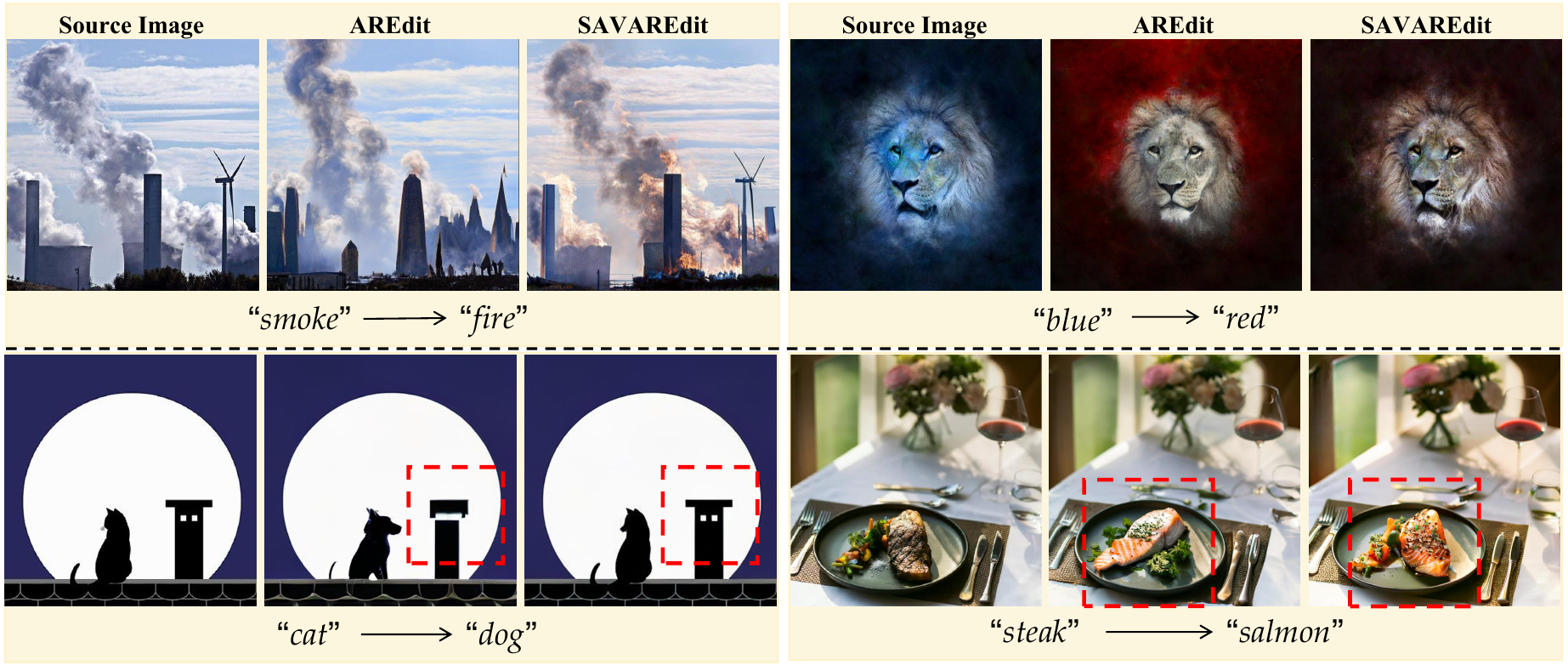}
  \captionof{figure}{\textbf{SAVAREdit for Text-Guided Image Editing.}
  Compared with AREdit \cite{wang2025training}, our method maintains consistent spatial structures with both local and global edits.}
  \label{fig:teasure}
  \vspace{-3mm}
\end{strip}

\input{sec/0_abstract}
\input{sec/1_intro}

\input{sec/2_relatedwork}

\input{sec/3_method}
\input{sec/4_experiments}
\input{sec/5_conclusion}
{
    \small
    \bibliographystyle{ieeenat_fullname}
    \bibliography{main}
}


\end{document}

%% file: preamble.tex
\usepackage{algorithmic}
\usepackage{algorithm}
\usepackage{makecell}

%% file: sec/0_abstract.tex
\begin{abstract}
Visual autoregressive (VAR) models have recently emerged as a promising family of generative models, enabling a wide range of downstream vision tasks such as text-guided image editing. By shifting the editing paradigm from noise manipulation in diffusion-based methods to token-level operations, VAR-based approaches achieve better background preservation and significantly faster inference. However, existing VAR-based editing methods still face two key challenges: accurately localizing editable tokens and maintaining structural consistency in the edited results. In this work, we propose a novel text-guided image editing framework rooted in an analysis of intermediate feature distributions within VAR models. First, we introduce a coarse-to-fine token localization strategy that can refine editable regions, balancing editing fidelity and background preservation. Second, we analyze the intermediate representations of VAR models and identify structure-related features, by which we design a simple yet effective feature injection mechanism to enhance structural consistency between the edited and source images. Third, we develop a reinforcement learning–based adaptive feature injection scheme that automatically learns scale- and layer-specific injection ratios to jointly optimize editing fidelity and structure preservation.
Extensive experiments demonstrate that our method achieves superior structural consistency and editing quality compared with state-of-the-art approaches, across both local and global editing scenarios.
\end{abstract}

%% file: sec/1_intro.tex
\section{Introduction}
\label{sec:intro}

Recent advances in text-to-image (T2I) generative models \cite{Gu2022Vector,Ho2020Denoising,Podell2023SDXL,Rombach2022HighResolution,Lipman2023FlowMatching,Liu2023FlowStraightFast,Song2023ScoreBasedGenerative} have led to a wide range of real-world applications \cite{Xia2025Consistent,Ruiz2023Dreambooth,Brooks2023Instructpix2pix,Li2024GetWhatYouWant,Chung2024StyleInjection,Chai2023Stablevideo,Meng2023SDEdit}, enabling users to create and manipulate visual content with unprecedented flexibility. Among them, text-guided image editing \cite{Brooks2023Instructpix2pix,Feng2025Dit4Edit,Brack2024LeditsPlusPlus,Hertz2023PromptToPrompt,Kawar2023Imagic} has emerged as a key technique that allows intuitive modification of existing images according to natural language descriptions, supporting diverse applications such as artistic creation, image enhancement, and interactive visual design.

Most existing text-guided image editing approaches are built upon pretrained diffusion models \cite{Hertz2023PromptToPrompt,Kawar2023Imagic,Brooks2023Instructpix2pix}. Although these works have achieved remarkable progress, the inherent inefficiency of diffusion models and their difficulty in accurately localizing the editing regions limit their applicability in real-world editing scenarios. Recently, a new family of text-to-image generative models, referred to as visual autoregressive (VAR) models, has emerged and shown strong potential across a variety of downstream vision tasks \cite{Tian2024Visual,Han2025Infinity,Tang2025HART,Chung2025FineTuning}. Given their efficiency and generative flexibility, researchers have begun exploring text-guided image editing within the VAR framework. AREdit \cite{wang2025training} introduces the first training-free text-guided image editing framework built upon VAR models. It represents a revolutionary shift in image editing from diffusion-based noise-space manipulation to token-level operations, where the target region is re-generated and the background tokens are preserved through a token reassembly strategy. In addition, attention control is also incorporated to maintain structural consistency. This design yields superior background preservation and more than 10x inference acceleration \cite{wang2025training} compared with the diffusion-based counterparts. Despite their great success, however, 
VAR-based editing methods still face two fundamental challenges: 
(1) Inaccurate localization of editable tokens often leads to unintended modifications in non-edited regions. 
(2) For global editing, existing approaches struggle to maintain fine-grained structural consistency as shown in \cref{fig:teasure}.

To address these limitations, we propose two novel techniques that improve token reassembly and leverage structure-related intermediate representations, resulting in finer structural control and more faithful editing outcomes.
(1) We start by examining the token reassembly strategy, which lies at the core of VAR-based editing methods, and find that the classifier-free guidance (CFG) parameter critically influences the trade-off between editing precision and background consistency. In particular, a larger CFG emphasizes the edited region but weakens background stability, while a smaller one preserves the background at the expense of edit quality. Hence, CFG implicitly affects the localization of editing regions. Based on this insight, we propose a coarse-to-fine editing-token localization strategy that achieves better editing fidelity while maintaining background unchanged. 
(2) Existing methods \cite{wang2025training,Dao2025DiscreteNoiseInversion} often reuse tokens from early scales and apply cross-attention control to maintain the spatial layout of the image. However, these strategies struggle to preserve fine-grained structural details as shown in \cref{fig:teasure} and tend to compromise the flexibility of editing. Diffusion-based editing frameworks, such as Prompt-to-Prompt (P2P) \cite{Dong2023Prompt}, Plug-and-Play (PnP) \cite{Tumanyan2023Plug}, and MasaCtrl \cite{Cao2023Masactrl}, preserve structural consistency by incorporating intermediate representations into the editing process. This naturally raises a question:
can VAR models provide analogous intermediate representations that maintain structural priors while retaining semantic controllability?
Through a layer-wise dissection and diagnostic analysis of the VAR framework, we show that intermediate representations within VAR exhibit strong spatial correspondence to the generated images.
Building upon this insight, we propose a novel Feature Injection (FI) mechanism that reutilizes spatial-related representations encoded in VAR models to reinforce structural consistency during text-guided image editing. To this end, we introduce a simple and effective reinforcement learning–based Adaptive Feature Injection (AFI) method that automatically determines the optimal injection ratios across different scales and layers. 

The main contribution of our work is SAVAREdit, a novel structure-aware image editing framework based on VAR models. Extensive qualitative and quantitative comparisons with state-of-the-art methods demonstrate its superior performance in both structure preservation and editing fidelity. Specifically, our approach features the following aspects:


\begin{itemize}{

\item A coarse-to-fine token localization (CFTL) strategy is proposed to refine the editing regions, achieving improved background preservation while maintaining high editing fidelity.

\item We conduct a comprehensive analysis of intermediate representations in VAR models and provide new empirical insights into their internal spatial feature formation. To the best of our knowledge, this is the first study that systematically explores the spatial feature distribution within the VAR models. 

\item 
A Feature Injection (FI) mechanism is designed to reutilize spatial representations derived from VAR models, and an Adaptive Feature Injection (AFI) module extends FI by incorporating reinforcement learning to automatically adjust the injection weights, achieving adaptive and structure-preserving image editing.
}
\end{itemize}

%% file: sec/2_relatedwork.tex
\section{Related work}
\label{sec:intro}


\noindent\textbf{Image Editing Based on Diffusion Models.}
In recent years, text-to-image diffusion models such as Stable Diffusion (SD) \cite{Rombach2022HighResolution} and SDXL \cite{Podell2023SDXL} have achieved remarkable progress, largely driven by large-scale training datasets like LAION-400M \cite{Schuhmann2021LAION} and Conceptual-12M \cite{Changpinyo2021Conceptual}. Building on these generative models, a large body of text-guided image editing methods have been developed, enabling local modification of image content from textual descriptions without explicit region specification \cite{Brooks2023Instructpix2pix,Kawar2023Imagic,Dong2023Prompt}. Among them, training-free methods have drawn particular attention, as they require no additional training or external data, instead exploit the intrinsic capacity of pretrained diffusion models for localized editing. Specifically, P2P \cite{Dong2023Prompt} injects cross-attention into the editing branch to maintain subject consistency. PnP \cite{Tumanyan2023Plug} leverages intermediate decoder and self-attention features to better preserve structural integrity, while MasaCtrl \cite{Cao2023Masactrl} further introduces key–value injection from self-attention layers to enhance subject coherence. In addition, Wang et al. \cite{Liu2024Towards} conducted a detailed analysis of cross- and self-attention mechanisms in diffusion models, showing that self-attention primarily captures global structural information with limited semantic content. Together with diffusion inversion techniques \cite{Mokady2023Null,HubermanSpiegelglas2024Edit,Tang2024Locinv,Dong2023Prompt} that enable faithful reconstruction and manipulation of real images, these studies have deepened the understanding of diffusion-based editing. Inspired by these advances, in this work, we aim to extend these explorations to the VAR framework.

\noindent\textbf{VAR-Based Image Generation and Editing.}
Originating from natural language processing (NLP), autoregressive (AR) models have recently been extended to the domain of visual content generation \cite{Sun2024Autoregressive,Mu2025Editar,Sun2025Personalized}. In this context, visual AR models tokenize images into discrete sequences and employ transformers to model the dependencies among visual tokens. Currently, visual AR models can be broadly categorized into two mainstream architectures: Next Token Prediction (NTP) \cite{Sun2024Autoregressive,Mu2025Editar,Sun2025Personalized} and Next Scale Prediction (NSP) \cite{Tian2024Visual,Han2025Infinity,Tang2025HART}. NTP follows the traditional autoregressive paradigm, formulating image synthesis as a next-token prediction problem, whereas NSP—also referred to as visual autoregressive (VAR) models—predicts visual content progressively across scales rather than individual tokens. Driven by the strong generative capability of VAR models, several studies have explored their potential in image editing. ATM \cite{Hu2025Anchor} proposes a novel training-free AR-based editing method that performs multiple sampling rounds at each token generation step and selects the candidate token closest to an anchor token as the final output. AREdit \cite{wang2025training} introduces the first VAR-based editing framework built upon the NSP architecture, marking a paradigm shift from noise-space manipulation to token-level operations. VAREdit \cite{Mao2025Visual} presents an instruction-guided image editing approach that formulates editing as a controllable generation task by fine-tuning VAR models with paired training data. 

%% file: sec/3_method.tex
\FloatBarrier 
\begin{figure*}[t]
  \centering
  \includegraphics[width=\linewidth]{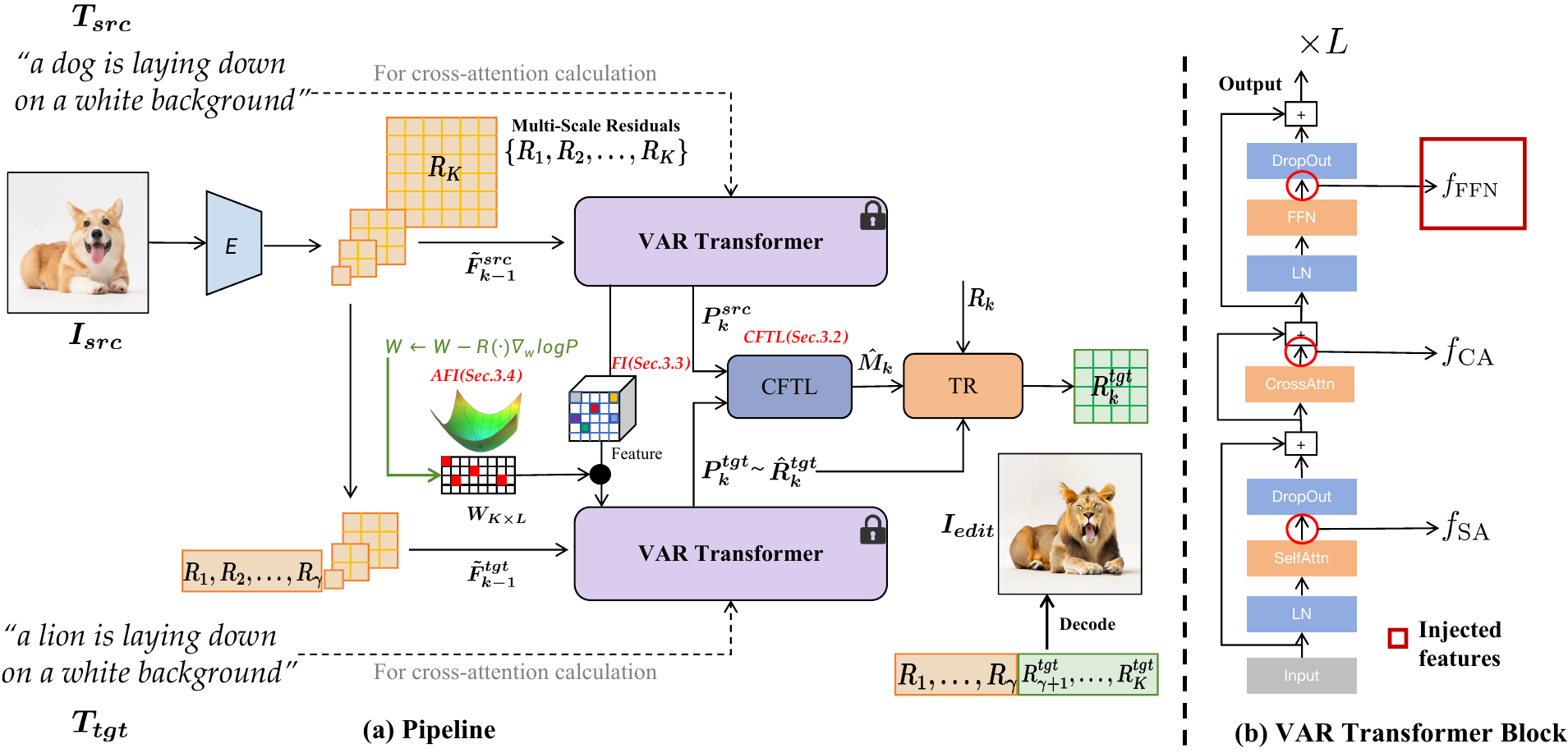}
  \caption{\textbf{Method overview.}
  Given a source image $I_{\text{src}}$ and its corresponding text description $T_{\text{src}}$, our framework generates the edited image $I_{\text{edit}}$ according to the target text $T_{\text{tgt}}$.
  The tokenizer $E$ encodes the input $I_{\text{src}}$ into multi-scale residuals $\{R_1,\ldots,R_K\}$.
  For clarity, we illustrate the inference pipeline at the $k$-th scale.
  The framework adopts a dual-branch architecture, where the source branch takes $\tilde{F}_{k-1}^{\text{src}}$ and the target branch takes $\tilde{F}_{k-1}^{\text{tgt}}$, producing probability distributions $P_k^{\text{src}}$ and $P_k^{\text{tgt}}$.
  The intermediate feature maps---illustrated in (b)---are selectively injected from the source branch into the target branch (\cref{subsec:SAVAR}) with learnable weights (\cref{subsec:AFI}).
  The predicted distributions $P_{k}^{\text{tgt}}$ are fed into the CFTL module (\cref{subsec:CFTL}) to obtain a refined editing mask $\hat{M}_k$, which is then used by token reassembly (TR) to produce the prediction $R_k^{\text{tgt}}$ at the $k$-th scale.
  Finally, multi-scale predictions $\{R_1,\ldots,R_{\gamma},R_{\gamma+1}^{\text{tgt}},\ldots,R_{K}^{\text{tgt}}\}$ are jointly decoded to produce the final edited image $I_{\text{edit}}$.
  Here, $\gamma$ denotes the number of source scales reused in the target branch.}
  \label{fig:pipeline}
\end{figure*}

\section{Method}
\label{sec:method}

In this section, we begin with a brief overview of the visual autoregressive (VAR) model in \cref{subsec:Preliminary}. Then, a coarse-to-fine token localization strategy is proposed for editing region refinement (\cref{subsec:CFTL}), followed by a structure-preserving method based on Feature Injection (FI), which is developed from an analysis of intermediate VAR representations (\cref{subsec:SAVAR}).
Finally, an Adaptive Feature Injection (AFI) mechanism driven by reinforcement learning is introduced (\cref{subsec:AFI}). our overall framework is presented in \cref{fig:pipeline}, and the detailed algorithmic pipeline is provided in Algorithm~1 of the supplementary material.

\subsection{Preliminary}
\label{subsec:Preliminary}
\noindent\textbf{Next Scale Prediction.} The visual autoagressive models using next-scale prediction architecture redefines the conventional paradigm of autoregressive (AR) modeling by transforming the token prediction process from a position-wise sequence generation task into a multi-scale token prediction problem.
Specifically, given an input image, a tokenizer $E$ first encodes it into a continuous feature map $F$. This feature map is then iteratively quantized into $K$ residual maps across multiple scales $R=(R_1, R_2, R_3, ..., R_K)$. Given previous token maps $R_{<k}$ and a condition $c$, the $k-th$ scale tokens $r_k$ are computed as:
\begin{equation}
p(R)=\prod_{k=1}^K p\left(R_k \mid R_{<k}, c\right),
\label{eq:nsp2}
\end{equation}
\noindent Then, we can progressively approximate the continuous feature $F$ as:
\begin{equation}
F_k=\sum_{i=1}^{k} \operatorname{upsample}\left(\operatorname{Lookup}\left(R_i\right)\right)
\label{eq:nsp1}
\end{equation}
\noindent where Lookup(·) is a lookup table mapping each token $R_k$ to its closest entry in a learnable codebook, and upsample(·) is a bilinear upsampling operation. For each step that predicts $R_k$, a bilinear downsampled feature $\tilde{F}_{k-1}$ is used as input, which is obtained from $F_{k-1}$. Recently, the next-scale prediction framework has demonstrated its effectiveness for large-scale text-to-image generation. Infinity \cite{Han2025Infinity} adopts bit-wise quantization techniques \cite{YuLanguage}, significantly expanding the available codeword space. Specifically, it introduces an additional dimension $d$ to $R_k$, extending it from a two-dimensional $(h_k, w_k)$ representation to $(h_k, w_k, d)$, 
where $h_k$ and $w_k$ represents the height and width of $k-th$ scale, $d$ denotes the number of bits assigned to each spatial position. 
Therefore, the size of the codebook becomes $2^d$, which increases exponentially with the bit dimension $d$. As a result, it achieves text fidelity and aesthetic quality comparable to state-of-the-art diffusion-based methods.


\subsection{Coarse-to-fine token localization}
\label{subsec:CFTL}

\noindent\textbf{Review of Token Reassembly (TR).} Different from diffusion-based editing methods that manipulate noise latents, VAR-based editing \cite{wang2025training} focuses on token-level operations, which identifies editable tokens that require modification while preserving the rest. It achieves this via a token reassemblely (TR) strategy that generates the editing binary mask according to the change of probability distributions predicted by the VAR model when conditioned on the source and target prompt, respectively. It is usually defined by:
\begin{equation}
\begin{aligned}
P^{\text{src}}_k &= G_\theta(\tilde{F}_{k-1}^{src}, \Psi(T_{\text{src}})), \\
P^{\text{tgt}}_k &= G_\theta(\tilde{F}_{k-1}^{tgt}, \Psi(T_{\text{tgt}}))
\end{aligned}
\label{eq:predict}
\end{equation}
\begin{equation}
M_k = \left[ ( P^{src}_k[\ldots, R_k] - P_k^{tgt}[\ldots, R_k] ) > \tau \right]
\label{eq:deltaP}
\end{equation}
\noindent where $G_\theta$ denotes the VAR model parameterized by $\theta$, $\Psi$ represents the text encoder such as Flan-T5 \cite{Chung2024Scaling}, $M_k$ denotes the binary mask at the $k$-th scale, $P_k^{src}$ represents the probability distribution predicted by the VAR model conditioned on the source prompt, $P_k^{tgt}$ corresponds to the distribution predicted under the target prompt, and $\tau$ is a threshold hyperparameter. After obtaining the mask, the tokens can be reassembled as shown according to:

\begin{equation}
R_k^{tgt} = M_k \odot \hat{R}_k^{tgt} + (1 - M_k) \odot R_k
\label{eq:token-reassemble}
\end{equation}

\noindent where $\hat{R}_k^{tgt}$ represents the tokens sampled from the distribution $P_k^{tgt}$, and ${R}_k^{tgt}$ denotes the final reassembled tokens at the $k$-th scale.
Consequently, the effectiveness of token-based editing fundamentally depends on accurately localizing the editable regions, i.e., obtaining a precise binary mask ${M}_k$ to guide token reconstruction.

\begin{figure}[!t]
  \centering
   \includegraphics[width=1.0\linewidth]{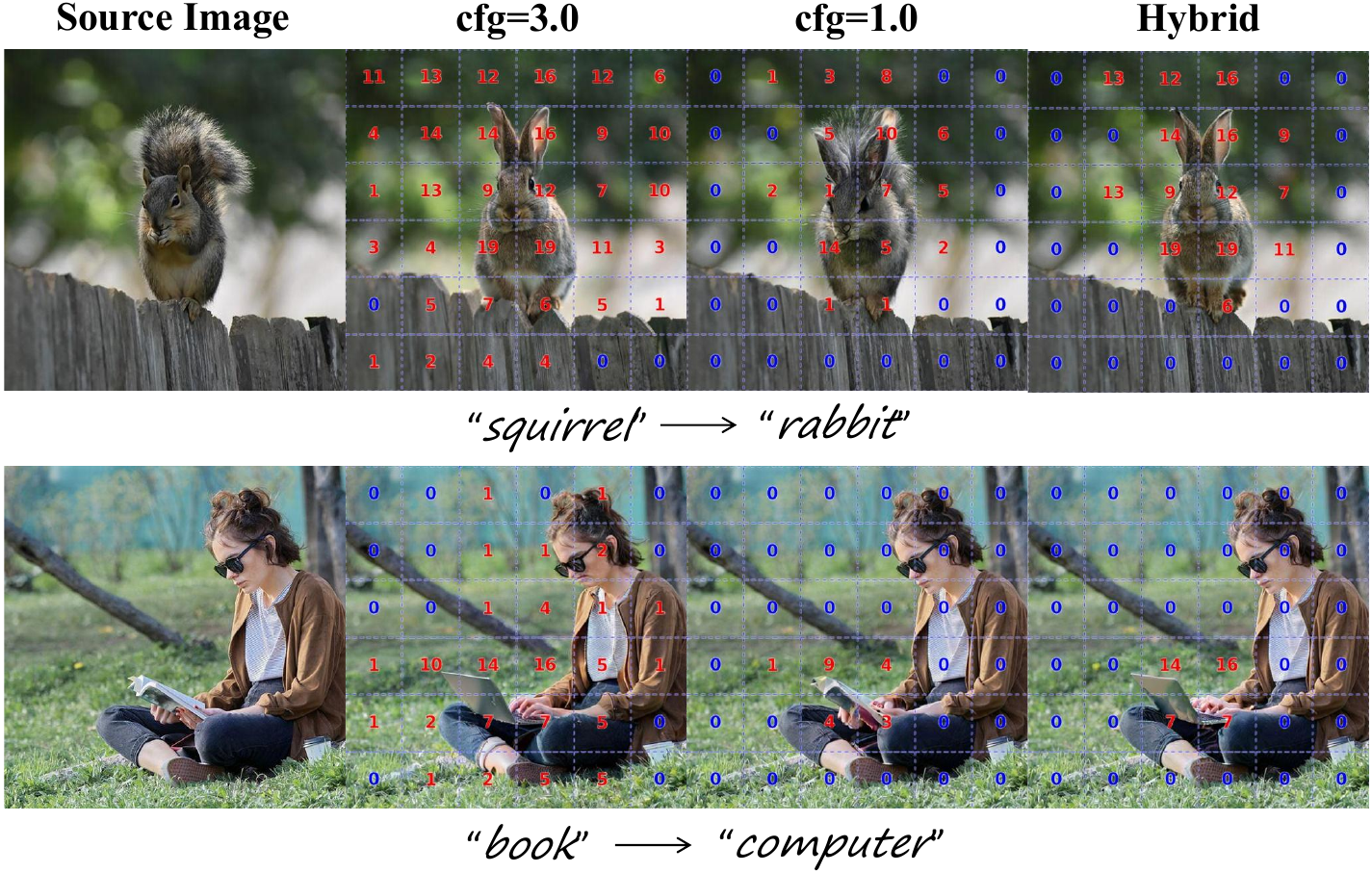}

   \caption{\textbf{Effect of CFG on background preservation and editing fidelity.} A higher CFG yields better editing fidelity but weaker background preservation, 
  whereas a lower CFG produces the opposite effect. 
  Our hybrid method achieves a good balance. 
  The number in each grid indicates the amount of tokens that need to be replaced at each position.}
   \label{fig:cfg}
\end{figure}

\noindent\textbf{Problem Analysis and Solution.} It is observed  that the accuracy of ${M}_k$ is highly sensitive to the setting of the classifier-free guidance (CFG) parameter. As shown in \cref{fig:cfg}, different CFG values lead to distinct trade-offs: a lower CFG tends to yield a more spatially precise mask but may reduce editing fidelity, whereas a higher CFG enhances visual fidelity at the cost of background consistency. It indicates that the CFG parameter implicitly affects the magnitude of probability variation, thereby influencing the localization accuracy of the editable regions.
To achieve a balance between background preservation and editing fidelity, a naive solution is to adopt a hybrid of low- and high-CFG inferences. Specifically, the first inference with a low CFG produces a coarse mask $M_{\text{coarse}} \in \mathbb{R}^{h_k \times w_k}$ that localizes editable tokens with spatial precision, whereas the second inference with a high CFG generates a fine-grained mask 
$\hat{M}_k \in \mathbb{R}^{h_k \times w_k \times d}$ that refines bit-level details.
Finally, the intersection of these two masks is taken to produce the final mask $\hat{M}_k$ as illustrated in \cref{eq:mask-intersection}, which enables more accurate localization of the editable regions as illustrated in the third column of \cref{fig:cfg}. 
\begin{equation}
\hat{M}_k = M^{\text{coarse}}_k \odot M^{\text{fine}}_k
\label{eq:mask-intersection}
\end{equation}
Nevertheless, performing two separate inference passes is obviously inefficient. 
To address this issue, we provide a theoretical analysis of this phenomenon as follows:
\begin{equation}
\begin{aligned}
P^{\text{src}}_k &= G_\theta(\tilde{F}_{k-1}^{src}, \emptyset) \\
&\quad + w \big(G_\theta(\tilde{F}_{k-1}, \Psi(T_{\text{src}})) 
- G_\theta(\tilde{F}_{k-1}, \emptyset)\big)
\label{eq:cfg-prob-src}
\end{aligned}
\end{equation}

\begin{equation}
\begin{aligned}
P^{\text{tgt}}_k &= G_\theta(\tilde{F}_{k-1}^{tgt}, \emptyset) \\
&\quad + w \big(G_\theta(\tilde{F}_{k-1}, \Psi(T_{\text{tgt}})) 
- G_\theta(\tilde{F}_{k-1}, \emptyset)\big)
\label{eq:cfg-prob-src}
\end{aligned}
\end{equation}

\begin{equation}
\begin{aligned}
\Delta P &= w \left(P^{\text{src}}_k - P^{\text{tgt}}_k\right) \\
         &= w \big(G_\theta(\tilde{F}_{k-1}, \Psi(T_{\text{src}})) 
         - G_\theta(\tilde{F}_{k-1}, \Psi(T_{\text{tgt}}))\big)
\label{eq:deltaP}
\end{aligned}
\end{equation}

\noindent where $w$ represents the classifier-free guidance (CFG) weight. As shown in \cref{eq:deltaP}, the CFG parameter essentially scales the probability variation $\Delta P$. Consequently, using a single threshold for binarization therefore results in inaccurate localization of the editing regions. Hence, performing dual inferences with different CFG values can be equivalently viewed as applying two adaptive thresholds for binarization as illustrated in the following equation.
\begin{equation}
M^{\text{coarse}}_k=[\Delta P > \tau_{coarse}],
M^{\text{fine}}_k=[\Delta P > \tau_{fine}]
\label{eq:mcoarsandmfine}
\end{equation}

\noindent where $\tau_{\text{coarse}}$ and $\tau_{\text{fine}}$ denote the threshold hyperparameters used for generating the coarse and fine-grained masks, respectively. It achieve the same refinement effect without introducing any additional computational burden to the pipeline.

\subsection{Structure-aware editing based on VAR}
\label{subsec:SAVAR}
\begin{figure}[htbp]
  \centering
   \includegraphics[width=1.0\linewidth]{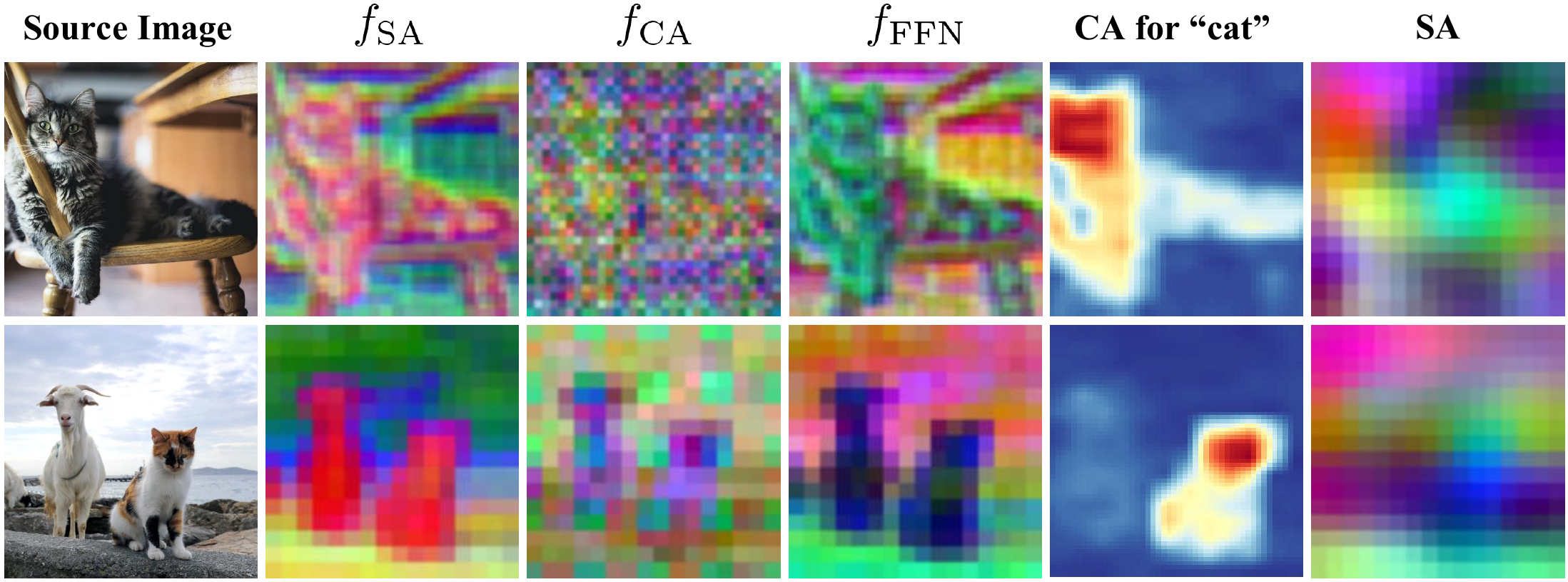}

   \caption{\textbf{VAR features and attention maps.} PCA visualization of intermediate representations and attention maps in a VAR block. Cross-attention (CA) offers rough object localization akin to diffusion models, but VAR self-attention (SA) does not display the spatial affinity structure typically observed in diffusion-based methods.}
   \label{fig:feature-analysis}
\end{figure}

\noindent \textbf{Spatial Feature}. In text-to-image generation, one can use descriptive text prompts to specify various scene and object proprieties, including those related to their shape, pose and scene layout, which together define the fundamental structural characteristics of an image. We opt to gain a better understanding of how such spatial information is internally encoded in VAR models. To this end, we perform a layer-wise dissection and visualize intermediate representations at different positions within transformer blocks using principal component analysis (PCA~\cite{pearson1901closestfit}), which allows us to better understand how spatial features are distributed within VAR models. Specifically, given a text prompt, we feed it into the VAR model and record the intermediate feature representations during the image generation process. 
We then apply PCA to reduce the dimensionality of these features for visualization. 
As shown in \cref{fig:feature-analysis}, the features extracted after the FFN ($f_{\mathrm{FFN}}$) and self-attention layers ($f_{\mathrm{SA}}$) clearly exhibit spatial information, 
a property analogous to the intermediate representations observed in the self-attention and decoder components of diffusion models as disscussed in \cite{Tumanyan2023Plug}. More details and analysis of the intermediate feature representations in VAR models are provided in the supplementary material.

\noindent\textbf{Feature Injection.} Based on this observation, we now turn back to the image editing task. Inspired by diffusion-based editing frameworks, we try to inject the cached intermediate features into the image generation process, aiming to enhance structural consistency during editing. 
Specifically, the source image $I_{\text{src}}$ is fed into the VAR model to extract intermediate features, denoted as $f^{\text{src}}_{s,l}$, where $s$ and $l$ index the scale and layer, respectively. We use $f^{\text{src}}_{s,l}$ as a unified notation for both $f_{\mathrm{FFN}}$ and $f_{\mathrm{SA}}$. Meanwhile, the target prompt $T_{\text{tgt}}$ is fed into the VAR model to generate the edited image $I_{\text{edited}}$, during the $k$-th scale prediction, the target features ${f^{\text{tgt}}_{s,l}}$ are overridden with ${f^{\text{src}}_{s,l}}$, as illustrated in the following equation.
\begin{equation}
R^{tgt}_{k} \sim G_\theta(\tilde{F}^{tgt}_{k-1}, \Psi(T_{\text{tgt}});\{f^{src}_{s,l}\})
\label{eq:fea-inj}
\end{equation}
\noindent where $\sim$ denotes sampling operation, and $R^{tgt}_{k}$ is drawn from the predicted probability distribution of $G_\theta(\tilde{F}^{tgt}_{k-1}, \Psi(T_{\text{tgt}}); \{f^{src}_{s,l}\})$ at the $k-th$ scale. It is observed that injecting features from the $f_{\mathrm{FFN}}$ yields better structural preservation than injecting those from the $f_{\mathrm{SA}}$. Therefore, unless otherwise specified, feature injection in the following sections refers to the injection of $f_{\mathrm{FFN}}$. More details about feature injection can be found in the supplementary material.

\begin{figure}[htbp]
  \centering
   \includegraphics[width=1.0\linewidth]{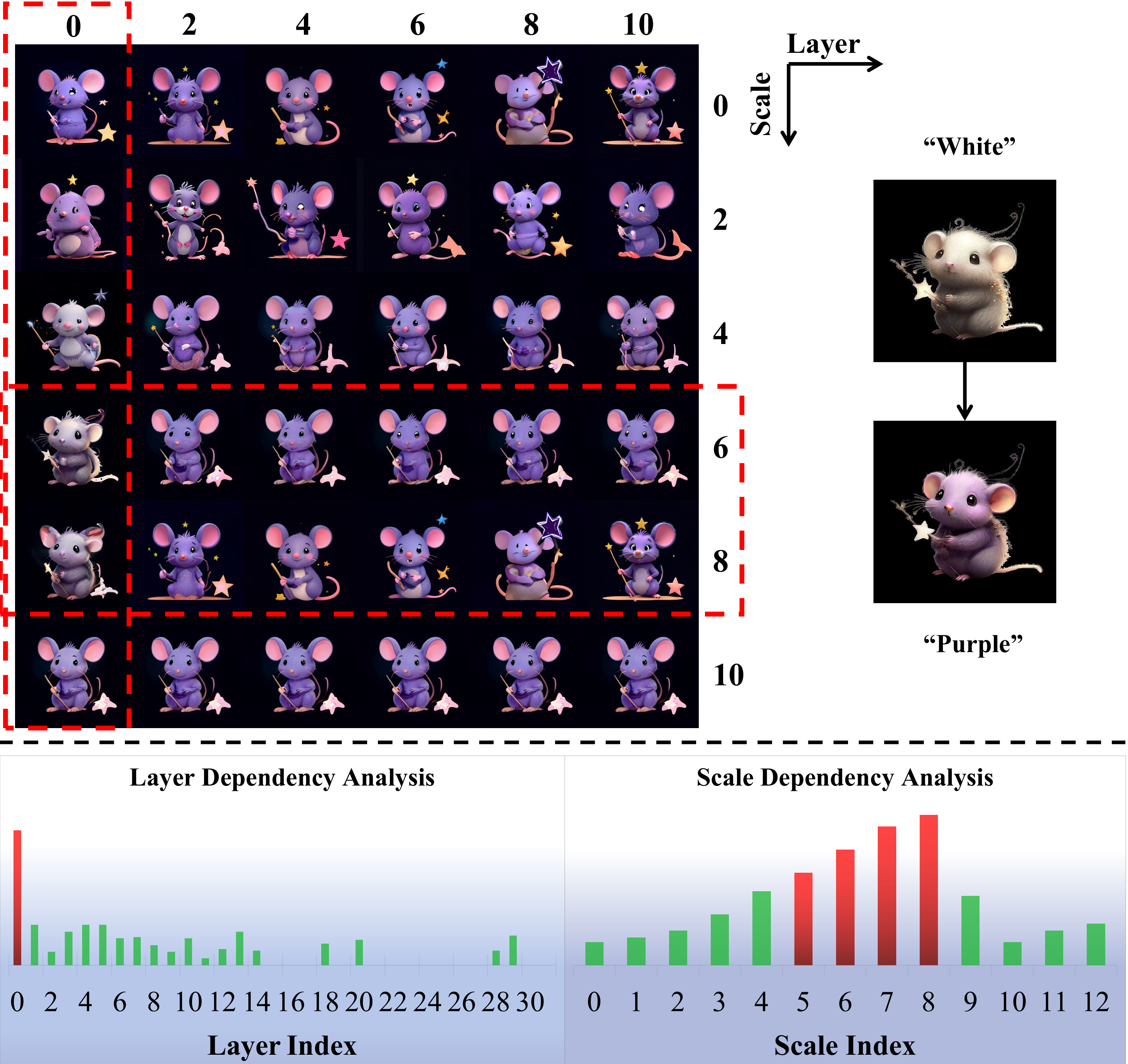}

   \caption{\textbf{Dependency analysis for feature injection.} \textbf{Top:} Experimental results of injecting features at different scales and layers, where the numbers denote layer and scale indices. Injecting features at the 0-th layer and at scales 5--8 leads to markedly better structural preservation. \textbf{Bottom:} Feature injection ratios after genetic algorithm optimization.}
   \label{fig:Dependency-analysis}
\end{figure}

\noindent\textbf{Dependency Analysis.} Through preliminary experiments, It is found that feature injection shows strong potential for preserving image structure. It raises another key question: which scales and layers in VAR models are most correlated with structural formation, and does injecting features at these locations improve editing quality? 
To explore this question, we formulate it as an optimization problem over a binary weight matrix $W_{K \times L}$, where each entry $\{0,1\}$ indicates whether cached features are injected at a given scale–layer pair. 
We employ a simple genetic algorithm to optimize $W$, with the fitness function defined by
\begin{equation}
\mathcal{F} = \lambda_{\text{CLIP}} \cdot \text{CLIP}(I_{\text{edit}}, T_{\text{tar}})
           + \lambda_{\text{SSIM}} \cdot \text{SSIM}(I_{\text{edit}}, I_{\text{src}})
\label{eq:fitness}
\end{equation}
where $\lambda_{\text{CLIP}}$ and $\lambda_{\text{SSIM}}$ balance the contributions of semantic fidelity and structural preservation, and $\text{CLIP}(\cdot)$ \cite{Radford2021CLIP} and $\text{SSIM}(\cdot)$ \cite{Wang2004SSIM} represent clip score and ssim score, respectively. We apply this optimization to multiple examples (approximately 100 in total) and then statistically analyze the injection ratio for each layer and scale based on the learned $W$. As shown in \cref{fig:Dependency-analysis}, the $0$-th layer and the scales from 5 to 8 exhibit relatively high injection ratios, indicating their crucial roles in spatial layout formation within VAR models. To our knowledge, these observations shed light on how spatial features are organized within VAR architectures.
Furthermore, we conduct layer-wise feature injection experiments as shown in \cref{fig:Dependency-analysis} (Top) to provide more direct evidences.
\subsection{Adaptive feature injection}
\label{subsec:AFI}
In the previous subsection, we treat the weight matrix $W$ as a binary variable. 
However, this discrete formulation may not yield an optimal solution, as the ideal injection ratio at each scale–layer position could vary continuously. To address this issue, we reformulate $W$ as a continuous matrix and attempt to optimize it via gradient descent. Thus, the feature overriding operation can be reformulated as a blending operation as follows:
\begin{equation}
\hat{f}_{s,l} = w_{s,l} \times f^{src}_{s,l} + (1-w_{s,l}) \times f_{s,l}^{tgt}
\label{eq:feature-blend}
\end{equation}
Unfortunately, the image generation process in VAR models involves a sampling operation, a typical non-differentiable component that disrupts the computational graph and makes conventional gradient-based optimization infeasible. To overcome this limitation, we adopt the policy gradient method \cite{williams1992simple,Sutton1999Policy,schulman2017proximal} from reinforcement learning. Specifically, $W$ is fed into the VAR model to produce a predicted probability distribution, from which edited images are sampled. Each generated image is evaluated using a reward function to compute its corresponding reward, 
and the policy gradient for $W$ is then calculated as shown in the following equation. 
\begin{equation}
\nabla_{W} J(W)
= \frac{1}{N}\sum_{i=1}^{N}
R\!\left(I_{\text{edit}}^{(i)}, I_{src}, T_{tgt}\right)\;
\nabla_{W} \log P_{W}
\label{eq:pg-minibatch}
\end{equation}
\noindent where $R(\cdot)$ denotes the reward of the generated image, consistent with the fitness function defined in \cref{eq:fitness}, 
and $P_{W}$ represents the probability distribution predicted by the VAR model conditioned on $W$. Here, $N$ denotes the mini\text{-}batch size used to reduce the sampling variance of the policy gradient estimation. To accelerate convergence and guide $W$ rapidly toward sensitive regions of the optimization landscape, we use a warm-up stage based on Simultaneous Perturbation Stochastic Approximation (SPSA) \cite{Spall2002Multivariate}. More details are presented in the supplementary material.


%% file: sec/4_experiments.tex
\begin{figure*}[htbp]
   \includegraphics[width=1.0\linewidth]{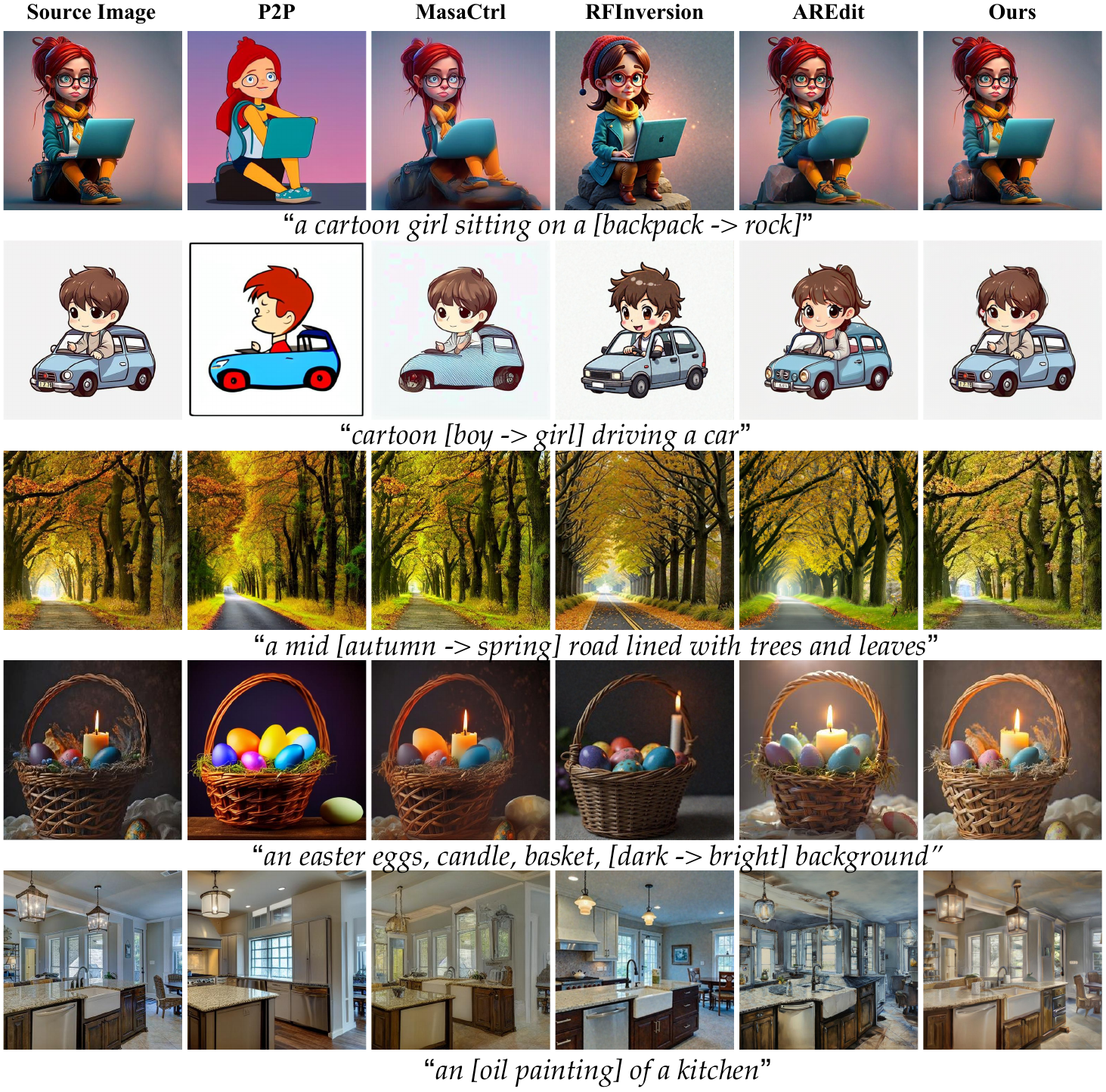}
    \caption{\textbf{Qualitative comparison with baselines.} Visual results of text-guided image editing from P2P~\cite{Dong2023Prompt}, MasaCtrl~\cite{Cao2023Masactrl}, RFInversion~\cite{Rout2025Semantic}, AREdit~\cite{wang2025training} and SAVAREdit (Ours). The original images and corresponding source/target prompts are provided. Our method achieves superior detail preservation in non-edited regions for local edits, and better structural consistency in global editing scenarios.}
   \label{fig:Qualitative-Comparison}
\end{figure*}

\section{Experiments}
\label{sec:experiments}

\subsection{Experimental setup}
\textbf{Baselines.} We implemented our method and compared it against several state-of-the-art text-guided image editing baselines, 
including both diffusion-based and VAR-based approaches: 
P2P \cite{Dong2023Prompt}, Pix2Pix-Zero \cite{Parmar2023Zero}, MasaCtrl \cite{Cao2023Masactrl}, 
PnP \cite{Tumanyan2023Plug}, PnP-DirInv \cite{Ju2024PnP}, LEdits++ \cite{Brack2024LeditsPlusPlus}, 
RFInversion \cite{Rout2025Semantic}, and AREdit \cite{wang2025training}. 

\noindent\textbf{Dataset.}
We evaluate our approach on the PIE-Bench dataset \cite{Ju2024PnP}, 
a comprehensive benchmark designed for text-guided image editing that contains 700 images encompassing both real-world and synthetic scenes. 

\noindent\textbf{Evaluation Metrics.}
We quantitatively evaluate our method in terms of structural similarity, text–image alignment, and fidelity in non-edited regions.
Text–image consistency is measured by CLIP similarity \cite{Wu2021GODIVA}, while LPIPS \cite{Zhang2018Unreasonable}, SSIM \cite{Wang2004SSIM}, and PSNR \cite{Hore2010PSNR} jointly assess perceptual, structural, and pixel-level fidelity in unedited areas.
In addition, Structural Distance \cite{Tumanyan2022Splicing} based on DINO-ViT \cite{Caron2021Emerging} features is used to evaluate global structural consistency.

\noindent\textbf{Implementation Details.} We adopt Infinity-2B \cite{Han2025Infinity}, the latest VAR-based text-to-image foundation model, as our backbone. We set $\gamma=3$, $\tau_{\text{coarse}}=0.2$, and $\tau_{\text{fine}}=0.6$ for local editing tasks such as object replacement. When feature injection is enabled, these parameters are adjusted to $\gamma=0$, $\tau_{\text{coarse}}=0.1$, and $\tau_{\text{fine}}=0.3$ for global editing scenarios such as style transfer and background modification, and we set $\lambda_{\text{CLIP}}=3.0$ and $\lambda_{\text{SSIM}}=1.0$ to achieve a better trade-off between background preservation and editing fidelity. All experiments are conducted on a single NVIDIA L20 GPU.

\subsection{Comparison with state-of-the-art baselines}

We present qualitative comparisons with existing state-of-the-art methods in \cref{fig:Qualitative-Comparison}, 
where our approach demonstrates superior editing fidelity and structural consistency across a variety of challenging scenarios. More visual results are provided in the supplementary material.
\cref{tab:quantitative} presents the quantitative results, further confirming that our approach effectively preserves the original structure while ensuring edits closely align with the intended modifications.


\begin{table}[t]
\centering
\normalsize 
\setlength{\tabcolsep}{4pt} 
\renewcommand{\arraystretch}{1.05} 
\resizebox{\columnwidth}{!}{ 
\begin{tabular}{l|c|c|ccc|cc}
\toprule
\textbf{Method} & \textbf{Base Model} &
\multicolumn{1}{c|}{\textbf{Structure}} &
\multicolumn{3}{c|}{\textbf{Background Preservation}} &
\multicolumn{2}{c}{\textbf{CLIP Similarity}$\uparrow$} \\
 &  & Distance$\downarrow$ & PSNR$\uparrow$ & SSIM$\uparrow$ & LPIPS$\downarrow$ & Whole & Edited \\
\midrule
P2P \cite{Dong2023Prompt} & diffusion & 0.0694 & 17.87 & 0.7114 & 0.2088 & 25.01 & 22.44 \\
Pix2Pix-Zero \cite{Parmar2023Zero} & diffusion & 0.0617 & 20.44 & 0.7467 & 0.1722 & 22.80 & 20.54 \\
MasaCtrl \cite{Cao2023Masactrl} & diffusion & 0.0284 & 22.17 & 0.7967 & 0.1066 & 23.96 & 21.16 \\
PnP \cite{Tumanyan2023Plug} & diffusion & 0.0282 & 22.28 & 0.7905 & 0.1134 & 25.41 & 22.55 \\
PnP-DirInv. \cite{Ju2024PnP} & diffusion & 0.0243 & 22.46 & 0.7968 & 0.1061 & 25.41 & 22.62 \\
LEdits++ \cite{Brack2024LeditsPlusPlus} & diffusion & 0.0431 & 19.64 & 0.7767 & 0.1334 & \textbf{26.42} & \textbf{23.37} \\
\midrule
RF-Inversion \cite{Rout2025Semantic} & flow & 0.0406 & 20.82 & 0.7192 & 0.1900 & 25.20 & 22.11 \\
\midrule
AREdit \cite{wang2025training} & VAR & 0.0305 & 24.19 & 0.8370 & 0.0870 & 25.42 & 22.77 \\
Ours & VAR & \textbf{0.0225} & \textbf{25.73} & \textbf{0.8521} & \textbf{0.0636} & 25.11 & 22.71 \\
\bottomrule
\end{tabular}}
\vspace{1mm}
\caption{\textbf{Quantitative comparison with baselines.}
Our method achieves comparable CLIP similarity to other state-of-the-art methods while demonstrating superior structure preservation.}
\label{tab:quantitative}
\end{table}

\subsection{Ablation study}
\label{sec:abalation}
\begin{figure}[htbp]
  \centering
   \includegraphics[width=1.0\linewidth]{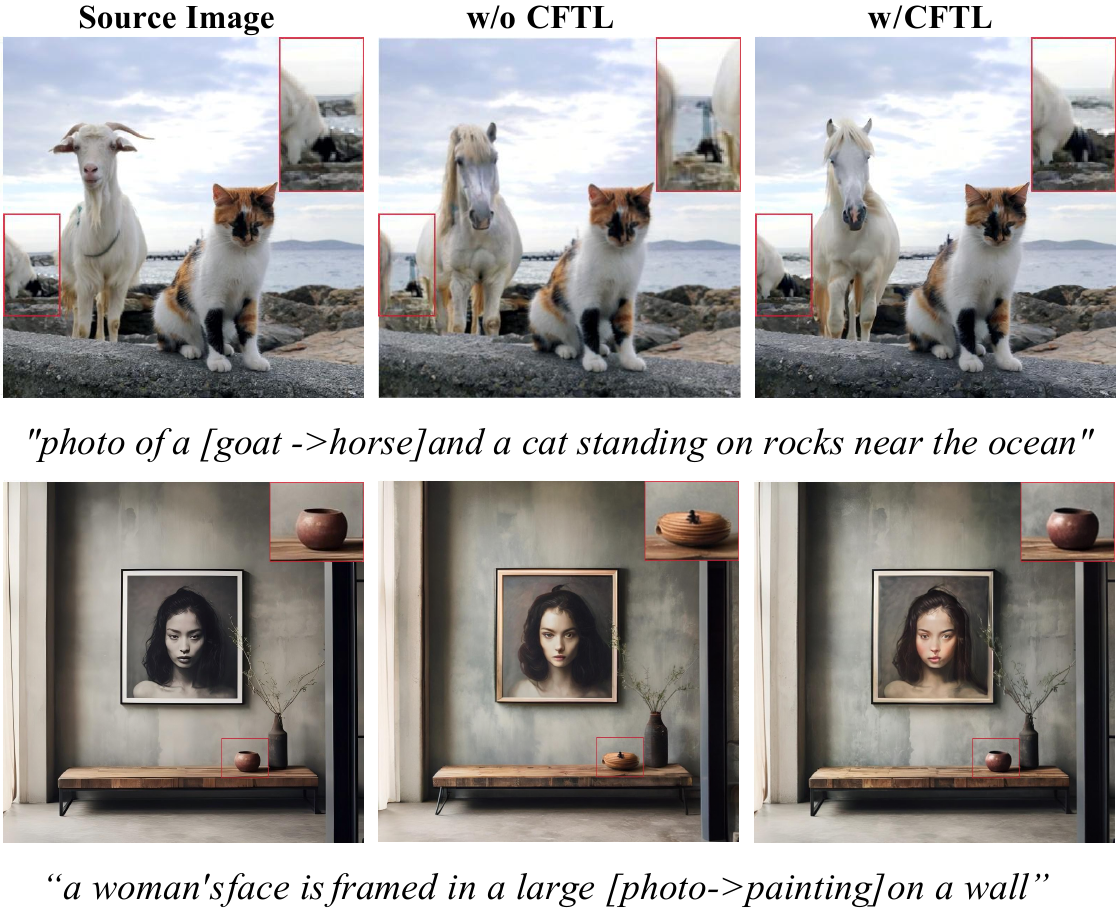}

   \caption{\textbf{Ablation study on CFTL.} CFTL enables more precise localization of editable regions, thereby achieving better background preservation while maintaining high editing fidelity.}
   \label{fig:ablation-cftl}
\end{figure}

\begin{figure}[htbp]
   \includegraphics[width=1.0\linewidth]{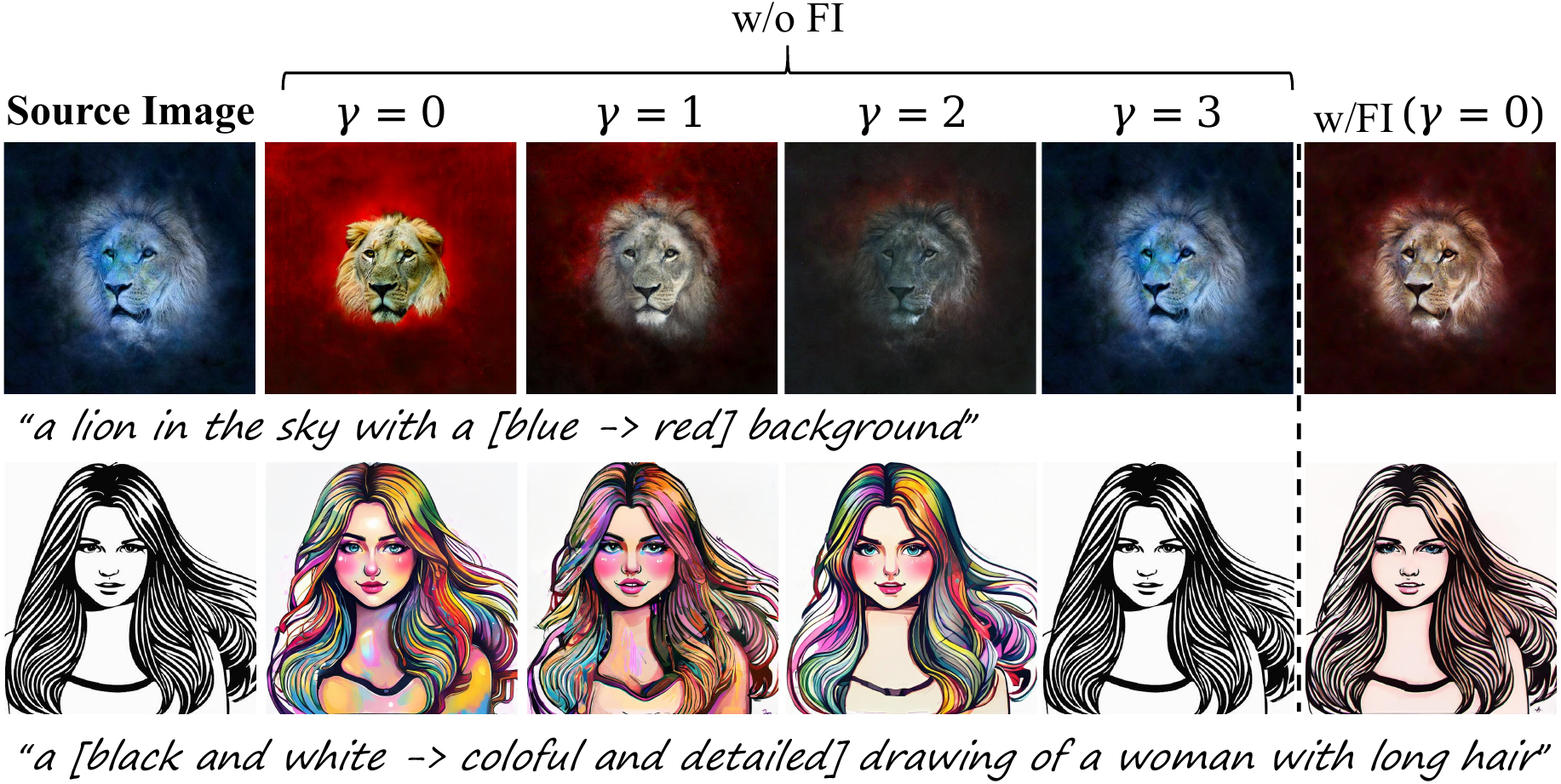}

   \caption{\textbf{Ablation study on FI.} FI exhibits strong capability in preserving image structures even without reusing any tokens ($\gamma=0$). Moreover, compared with the token-reuse strategy ($\gamma > 0$), FI provides greater flexibility and achieves a better balance between structural consistency and editing fidelity.}
   \label{fig:ablation-FI}
\end{figure}

\begin{figure}[htbp]
   \includegraphics[width=1.0\linewidth]{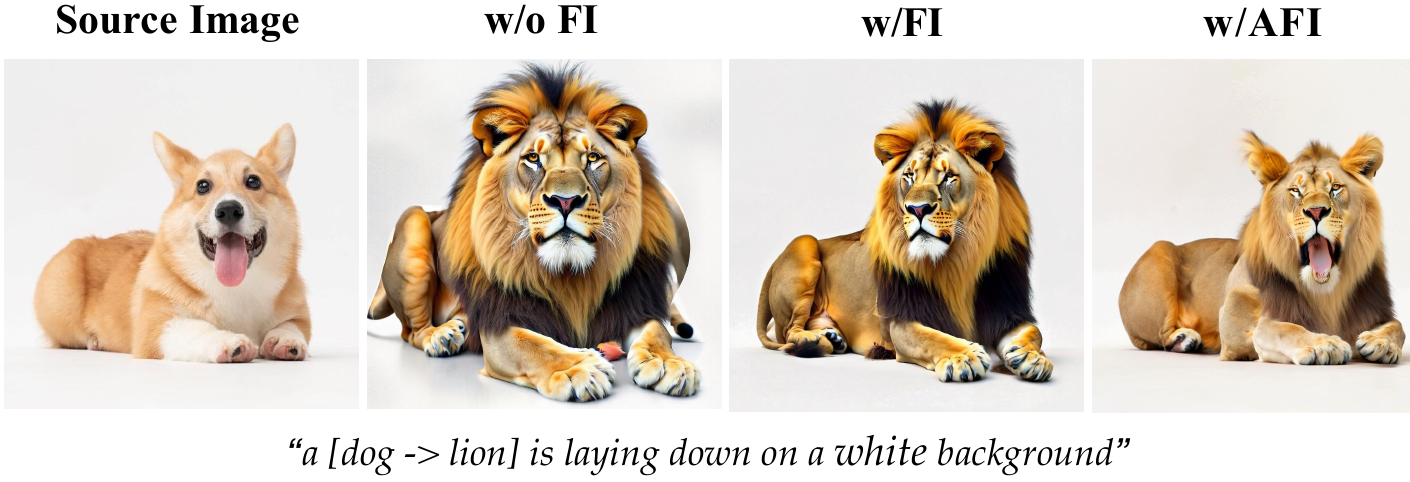}
   \caption{\textbf{Ablation study on AFI.} AFI achieves a better balance between structural consistency and editing fidelity.}
   \label{fig:ablation-AFI}
\end{figure}


\noindent\textbf{CFTL.} As shown in \cref{fig:cfg} and \cref{fig:ablation-cftl}, incorporating CFTL enables more precise localization of editable regions. It effectively distinguishes editable and non-editable areas, leading to clearer edit boundaries and better background preservation.


\noindent\textbf{FI \& AFI.} As shown in \cref{fig:ablation-FI}, FI effectively maintains image structure without relying on token reuse, achieving a better trade-off between structural consistency and editing fidelity. Ablation study on AFI is presented in \cref{fig:ablation-AFI}, it demonstrates that AFI provides greater flexibility in balancing structure preservation and editing fidelity. More qualitative and quantitative ablation results are presented in the supplementary material.



%% file: sec/5_conclusion.tex
\section{Conclusion}
\label{sec:conclu}



This paper presents a novel text-guided image editing framework 
rooted in new insights into the internal representations of pre-trained text-to-image Visual autoregressive (VAR) models. 
Our approach performs lightweight manipulations on intermediate representations to achieve a more effective balance between structural preservation and editing fidelity. 
Extensive experiments demonstrate that our method consistently outperforms existing diffusion- and VAR-based baselines in both structure preservation and editing fieldty.

As the future work, we plan to extend our approach to more powerful generative backbones 
and further explore hybrid diffusion–autoregressive strategies for improved visual realism and controllability.